\documentclass[a4paper, twocolumn]{IEEEtran}
\usepackage{textcomp}
\usepackage{xcolor} 
\usepackage{amsmath}
 
\usepackage{amssymb}
\usepackage{graphicx}
\usepackage{booktabs}
\usepackage{algorithm}
\usepackage{algpseudocode}
\usepackage{caption}
\usepackage{comment}  
\usepackage{authorindex}
\usepackage{subcaption}
\usepackage{multirow} 
\usepackage{float} 
\usepackage{balance} 

\begin{document}
  
      
 \title{Recursive weight-sharing transformers for semiconductor thermo-mechanical reliability}
 


\author{\IEEEauthorblockA{Kart-leong Lim  \\
 Institute of Microelectronics (IME), \\
 \textit{Agency for Science, Technology and Research (A{*}STAR),} 
Singapore \\
 limkl@a-star.edu.sg}}

\maketitle

\begin{abstract} 
Transformer-based surrogate models are increasingly used to replace expensive first-principles simulation in engineering design. But conventional transformer architectures are often over parameterized for the small, low-dimensional datasets typical of engineering design spaces, where large simulation data is expensive to generate. Under these conditions, excess parameter capacity leads to overfitting rather than improved accuracy, while also incurring unnecessary memory and compute overhead — this motivates a shift towards architectures that focus on  \textbf{additional compute} rather than \textbf{additional learnable parameters}. This paper presents a hardware-aware evaluation of three recursive transformer paradigms for surrogate thermo-mechanical analysis of advanced packages: a)Tiny Recursive Model, b) our proposed Depth Recursive transformer, c) and a simple recursive transformer. We systematically compare their predictive performance (Recall@K, Mean Reciprocal Rank), parameter count, computational complexity (FLOPs), providing practical design guidelines for selecting recursive transformer architectures under resource-constrained scenarios. We validate this principle on two low-dimensional engineering prediction tasks: 1) thermo-mechanical reliability analysis of advanced semiconductor packages, where stress and warpage from thermal cycling must be evaluated repeatedly across a design-of-experiments sweep under costly finite element analysis (FEA). 2) Laplace PDE iterative numerical solver for capacitance field. Overall, recursive weight-sharing transformers provide an effective and generalizable trade-off between prediction accuracy, parameter efficiency, and computational cost for small-data engineering surrogate modeling — a regime for which conventional large-parameter transformers are poorly suited — as demonstrated on advanced package reliability prediction and capactior electrostatic field modeling. 

\end{abstract}


\section{Introduction} 
Transformer models have demonstrated strong predictive performance across a wide range of engineering applications, including electronic design automation (EDA) \cite{fan2024gtn}, scientific machine learning \cite{li2023transformer}, and digital twins \cite{zhang2026wind}. However, their increasing computational and memory requirements present significant challenges for deployment in hardware-constrained environments, such as on-chip inference, edge accelerators, and real-time design optimization \cite{saha2025vitedge,cioflan2024mcu}. Consequently, substantial research has focused on reducing transformer computation through improvements at the algorithmic, architectural, and system levels \cite{dao2022flashattention,fedus2021switch,dehghani2018universal,han2015pruning}.  
Thermo-mechanical simulation remains computationally expensive, motivating the use of transformer-based surrogate models for rapid reliability prediction. However, deploying such models within practical EDA workflows requires architectures with low computational complexity, compact memory footprints, and high inference throughput. These requirements have motivated a broad range of transformer compute reduction techniques and can be broadly categorized into seven complementary paradigms according to the aspect of computation they optimize. These include \textbf{Efficient Attention} \cite{katharopoulos2020performer,dao2022flashattention}, \textbf{Token Reduction} \cite{rao2021dynamicvit,bolya2022tome}, \textbf{Conditional Computation} \cite{fedus2021switch,du2021glam}, \textbf{Architectural Optimization} \cite{dehghani2018universal,lan2020albert,yang2024looped,trm2025}, \textbf{Model Compression} \cite{han2015pruning,hinton2015distilling,hu2022lora}, \textbf{Hardware Optimization} \cite{dao2023flashattention2,nvidia2023tensorrtllm}, and \textbf{ State Space Models} \cite{gu2022s4,gu2024mamba}. Although these approaches pursue the common objective of improving computational efficiency, they address different computational bottlenecks, ranging from reducing attention complexity and sequence length to redesigning network architectures or replacing attention-based models altogether.
Among these paradigms, this work focuses on \textbf{Architectural Optimization}, where computational efficiency is achieved by redesigning the transformer itself. In particular, recursive weight sharing enables deeper computation without a proportional increase in parameters \cite{dehghani2018universal,yang2024looped,trm2025}, making it particularly attractive for resource-constrained EDA applications. Although numerous approaches have been proposed to reduce transformer computation, their suitability depends on the target application. Efficient attention mechanisms primarily benefit long-sequence models where self-attention dominates the computational cost \cite{katharopoulos2020performer,dao2022flashattention}, while token reduction techniques rely on large token sets to achieve meaningful computational savings \cite{rao2021dynamicvit,bolya2022tome}. Conditional computation and Mixture-of-Experts architectures improve scalability for large foundation models but introduce additional routing complexity and memory overhead \cite{fedus2021switch,du2021glam,mixtral2024}. Model compression and hardware-specific optimizations are generally applied after model design or target specific execution platforms \cite{han2015pruning,hu2022lora,dao2023flashattention2}. In contrast, our application involves compact transformer surrogates operating on short input sequences under strict memory and computational constraints. Consequently, Architectural Optimization  provides the most suitable design strategy, enabling computation to be reduced intrinsically through recursive weight sharing while preserving a compact, hardware-friendly architecture \cite{dehghani2018universal,yang2024looped,trm2025}.

\begin{figure}
       \centering \includegraphics[width=.5\textwidth]{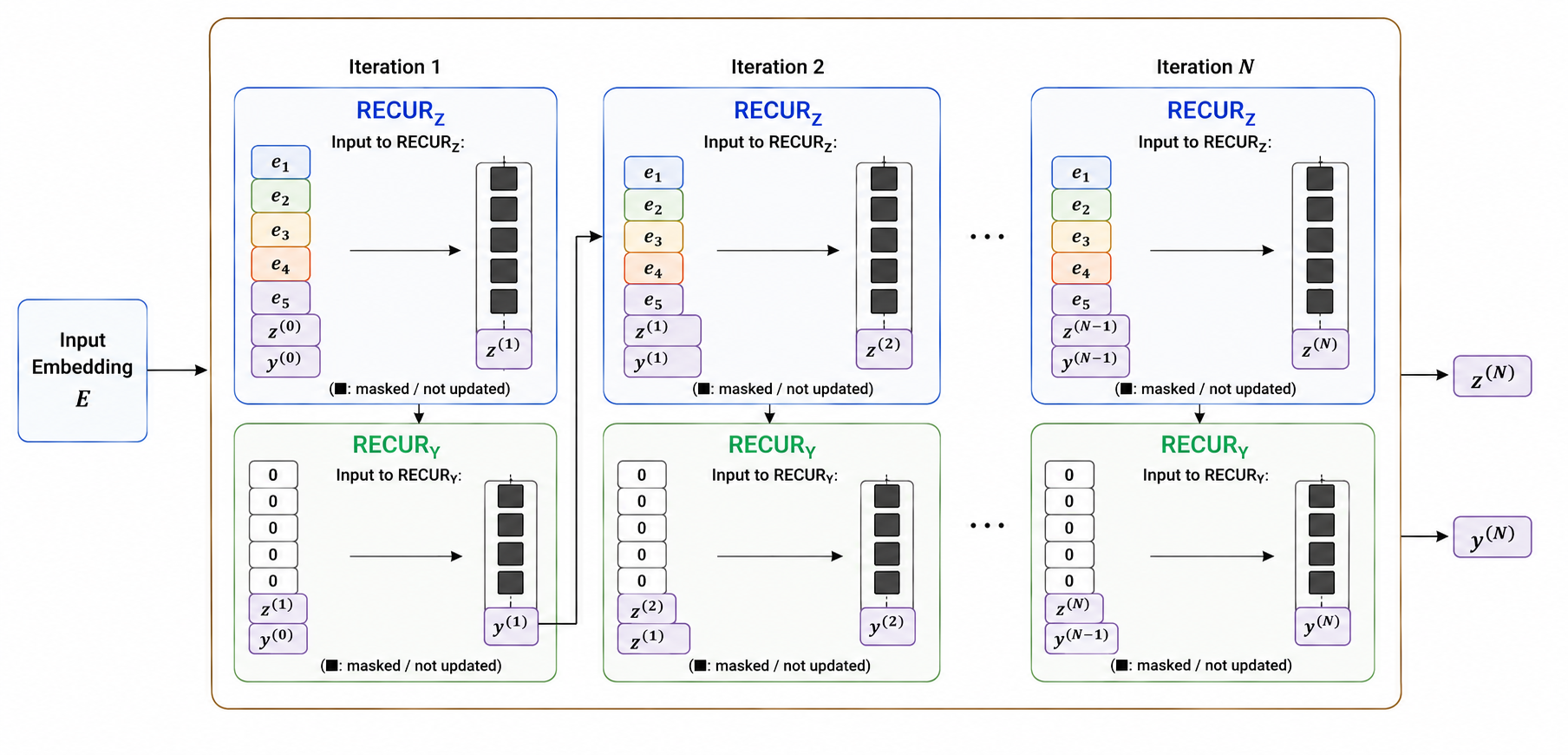} 
  \caption{Tiny Recursive model}
 \label{fig:Tiny Recursive model}
\end{figure}

\begin{figure}
        \centering \includegraphics[width=.5\textwidth]{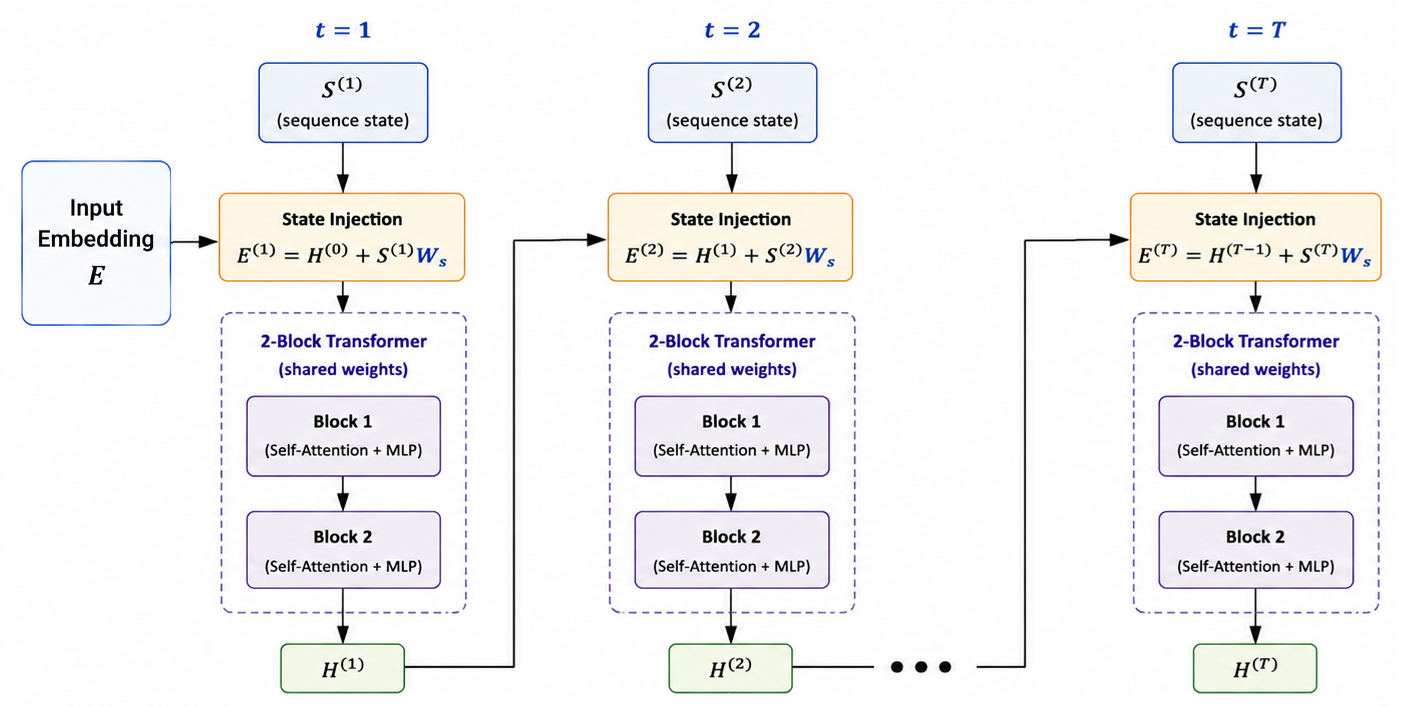} 
    \caption{Depth Recursive model}
    \label{fig:Depth Recursive model}
\end{figure}

\begin{figure}
       \centering \includegraphics[width=.5\textwidth]{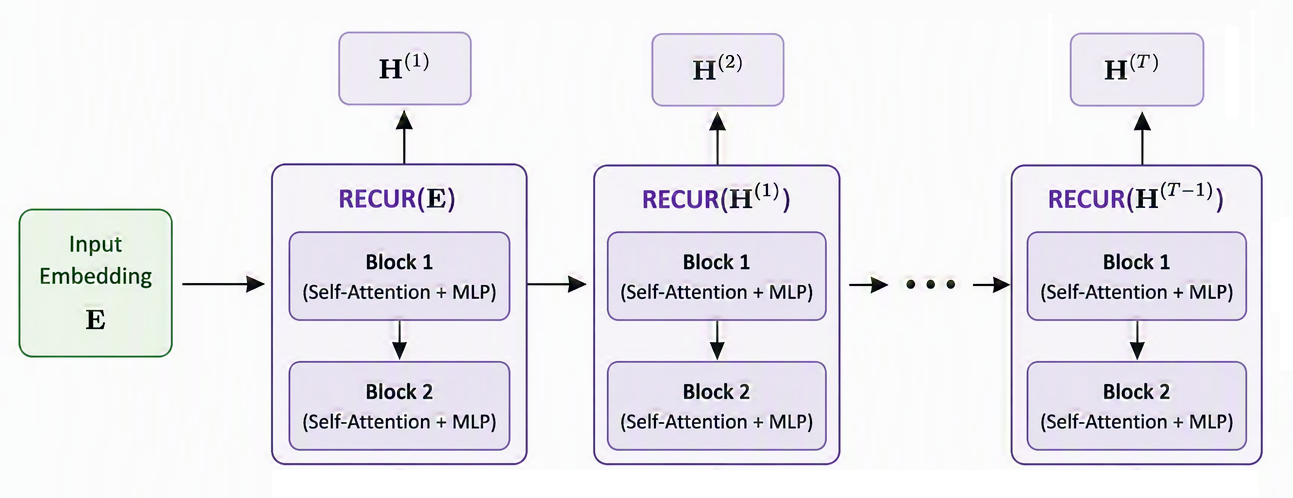} 
  \caption{Simple Recursive model}
   \label{fig:Simple Recursive model}
\end{figure}

\section{Recursive weight sharing transformers}
The main concept of recursive transformer is to treat transformer as a reusable template (e.g. $\operatorname{RECUR}$) throughout the model. Multiple replicas of RECUR are used in a model, but they all share the single set of weights, which is updated once during each training iteration. $\operatorname{RECUR}$ should be independent of models we choose to deploy in. 
The simplest configuration of $\operatorname{RECUR}$ is a one block transformer in Eqn.\ref{eq:shared_recur} which consists of Self-Attention $\operatorname{ATTN}(\cdot)$, Normalization layer $\operatorname{NORM}(\cdot)$, Feed-Forward Network (FFN). Conversely, if we want a deeper transformer, we can define $\operatorname{RECUR}$ using e.g. two transformer blocks. 
 
\begin{equation}
\begin{aligned}
\operatorname{RECUR} ={}&
\operatorname{ATTN}(\cdot \mid W_Q,W_K,W_V,W_O)
\\
&\rightarrow
\operatorname{ADD}_1+\operatorname{NORM}_1(\cdot \mid \gamma_1,\beta_1)
\\
&\rightarrow
\operatorname{FFN}(\cdot \mid W_1,W_2)
\\
&\rightarrow
\operatorname{ADD}_2+\operatorname{NORM}_2(\cdot \mid \gamma_2,\beta_2).
\end{aligned}
\label{eq:shared_recur}
\end{equation}

\subsection{Tiny Recursive Model}

The Tiny Recursive Model (TRM) uses weight sharing RECUR to cap the parameters while perform AI reasoning on Sudoku and puzzle tasks shown in Fig. \ref{fig:Tiny Recursive model}. Within each iteration, $\operatorname{RECUR}_{Z}$ updates the latent feature $\mathbf{z}$ by feeding the input embeddings $\mathbf{E}\in \mathbb{R}^{L \times d_{token}}=[\mathbf{e}_1,\ldots,\mathbf{e}_L]$--where $L$ is number of tokens and $d_{token}$ is the embedding dimension--along with the previous latent feature $\mathbf{z}^{(n-1)}$, and the previous output feature $\mathbf{y}^{(n-1)}$. The updated latent feature $\mathbf{z}^{(n)}$ is then passed to $\operatorname{RECUR}_{Y}$, which updates the output representation $\mathbf{y}^{(n)}$ while masking the input embeddings with zeros. Repeating this coupled update for $N$ times defines the $\operatorname{LATENT}$ operation in Eqn.\ref{eq:latent}. This progressively refine both $\mathbf{z}^{(n)}$ and $\mathbf{y}^{(n)}$. 
\begin{equation}
\begin{aligned}
\operatorname{LATENT}
\triangleq
\left(
\operatorname{RECUR}_{Y}
\circ
\operatorname{RECUR}_{Z}
\right)^{N} \\
\label{eq:latent}
%
\mathbf{z}^{(n)}=
\operatorname{RECUR_Z}
\!\left(
[\mathbf{e}_1,\ldots,\mathbf{e}_L,
\mathbf{z}^{(n-1)},
\mathbf{y}^{(n-1)}]
\right), \\
 %
\mathbf{y}^{(n)}=
\operatorname{RECUR_Y}
\!\left(
[\mathbf{0},\ldots,\mathbf{0},
\mathbf{z}^{(n)},
\mathbf{y}^{(n-1)}]
\right). 
\end{aligned}
\end{equation}
Finally to compute TRM (Eqn.\ref{eq:trm}), $\operatorname{LATENT}$ is executed for $T$ times and fed into another $\operatorname{RECUR}$. Increasing either the inner $N$ or the outer $T$ loops increases the computation cost (FLOPs) without increasing the parameters through weight sharing.   
\begin{equation}
\begin{split}
&\mathrm{TRM}\!\left(
\mathbf{E},
\mathbf{z}^{ini},
\mathbf{y}^{ini}
\right) 
\triangleq \operatorname{RECUR} \circ \left( \operatorname{LATENT} \right)^{T} \\
&= \operatorname{RECUR} \!\left(\mathbf{E},  \left[\left( \operatorname{RECUR}_{Y} \!\left( \mathbf{0}, \operatorname{RECUR}_{Z} \!\left( \mathbf{E}, \mathbf{z}, \mathbf{y} \right), \mathbf{y} \right) \right)^{N} \right]^{T} \right).
\end{split}
\label{eq:trm}
\end{equation}
 
\subsection{Depth Recursive Model}
From our domain expertise in advanced packaging, we observe that the dataset exhibits a gradually changing depth-like structure for each fixed design case. Motivated by this observation, we repurpose RECUR by treating sequential depth as a state input while fully retaining the weight sharing objective in Fig.\ref{fig:Depth Recursive model} We call this approach as the Depth Recursive Model (DEPTH). Inspired by Recurrent Neural Networks (RNNs), DEPTH adapts the notion of hidden states to weight-dependent sequential depth states within RECUR. Using the terminology of RNN for sequential modeling, in DEPTH (Eqn.\ref{eq:depth}) we define sequential state as $S^{(t)}$, initial hidden state as embedding input $\mathbf{E}=\mathbf{H}^{(0)}$, and sequential (or recursive) output $H^{(t+1)}$ for depth $t \in \{1, \dots, T\}$. 
\begin{equation}
\begin{array}{l}
\mathrm{DEPTH}\!\Big(
\mathbf{E}_{S^t \nsubseteq E},
\mathbf{S}^{(1)},\ldots,\mathbf{S}^{(T)}
\Big)
\triangleq
\left(
\operatorname{RECUR}
\circ
\left(\cdot+\mathbf{S}^{(t)}\mathbf{W}_s\right)
\right)^T \\[2mm]

=
\operatorname{RECUR}\!\left(
\mathbf{E}+\mathbf{S}^{(1)}\mathbf{W}_s
\right)
\rightarrow
\mathbf{H}^{(1)} \\

\rightarrow
\operatorname{RECUR}\!\left(
\mathbf{H}^{(1)}+\mathbf{S}^{(2)}\mathbf{W}_s
\right)
\rightarrow
\mathbf{H}^{(2)} \\

\cdots 
\rightarrow
\operatorname{RECUR}\!\left(
\mathbf{H}^{(T-1)}+\mathbf{S}^{(T)}\mathbf{W}_s
\right)
\rightarrow
\mathbf{H}^{(T)}.
\end{array}
\label{eq:depth}
\end{equation}

\subsection{Simple Recursive model}
The simple recursive model (SIMPLE) in Fig.\ref{fig:Simple Recursive model} is seen to repeatedly apply an identical recursive transformation. Using (Eqn.~\ref{eq:simple}), SIMPLE applies RECUR for $T$ recursive iterations to generate the output  $H^{(T)}$. 
\begin{equation}
\begin{aligned}
\operatorname{SIMPLE}(\mathbf{E}) &\triangleq \left( \operatorname{RECUR}\right)^T  \\
&= \operatorname{RECUR}(\mathbf{E}) \rightarrow \mathbf{H}^{(1)} \\
&\rightarrow \operatorname{RECUR}\left(\mathbf{H}^{(1)}\right) \rightarrow \mathbf{H}^{(2)} \\
&\dots \rightarrow \operatorname{RECUR}\left(\mathbf{H}^{(T-1)}\right) \rightarrow \mathbf{H}^{(T)}
\end{aligned}
\label{eq:simple}
\end{equation}

\section{Experimental Results}

\subsection{Baselines and proposed models}
To thoroughly evaluate the tracking and retrieval performance, we compare our approach against seven baseline configurations categorized across four distinct architectural paradigms:
\begin{itemize}
    \item \textbf{VANILLA:} Consists of \textbf{M1}, which serves as a baseline 1-Block Layer Normalization (LN) Transformer architecture.
    \item \textbf{SIMPLE:} Comprises \textbf{M2} and \textbf{M3} for Simple Recursive Model, representing a 2-Block Root Mean Square (RMS) recursive structure configured with recursive steps of $T=1$ and $T=3$, respectively.
    \item \textbf{TRM:} Represents the Tiny Recursive Model, where \textbf{M4} denotes a 1-Block-RMS variant incorporating a latent variable ($Z$), and \textbf{M5} denotes a 1-Block-LN variant operating over joint latent configurations ($Z,Y$). $T=5$ and $N=3$ were used.
    \item \textbf{DEPTH:} Comprises Depth Recursive Model, where \textbf{M6} specifies a 1-Block-RMS design and \textbf{M7} represents a 2-Block-RMS architecture. The depth is adapted to $T=16$ for both the Stress10K and Warpage10K dataset.
\end{itemize}

\subsection{Datasets description}
 \label{sec:dataset}

We evaluate recursive models across three datasets with partial DOE factorial cases, $N= factors^{levels}$---the Stress10k and Warpage10k datasets ($N=5^4 \times 16 =10{,}000$ pairs) generated by FEA simulation. Synthetic capacitor electrostatic field dataset generated by physics-informed neural network (PINN) ($N=22^2 \times 15 =7{,}265$ pairs). 

\textbf{Stress10k, Warpage10k} \cite{lim2025bdnet}: Advanced packaging design requires understanding how material properties and geometric configurations influence thermo-mechanical behavior across spatial and depth-wise dimensions. Design variables --epoxy molding compound (EMC) coefficient of thermal expansion (CTE) and elastic modulus (E), die sizes ($Size_{dies}$), inter-die spacing ($Gap_{dies}$), and position within the package depth ($S^{(t)}$) --determine the resulting stress and warpage responses obtained from finite element analysis (FEA).
\begin{equation} 
\begin{aligned}
\mathbf{x_{stress}}=
\begin{bmatrix}
\mathrm{EMC}_{\mathrm{CTE}},\;
\mathrm{EMC}_{E},\;
\mathrm{Size}_{\mathrm{dies}},\;
\mathrm{Gap}_{\mathrm{dies}},\;
S^{(t)}
\end{bmatrix}
\in\mathbb{R}^{5 \times 1}.
\\ 
\mathbf{I_{stress}} = \left\{ \mathbf{I}^{(1)},\mathbf{I}^{(2)},\dots,\mathbf{I}^{(16)} \right\}
\in\mathbb{R}^{13\times 13\times 16}
\end{aligned}
\end{equation} 
 \
\textbf{PINN} \cite{lim2023inverse}: The electrostatic field within an air-filled capacitor can be modeled by establishing collocation and boundary points across a two-dimensional grid. While the Laplace equation governs the internal collocation points, the system's boundary conditions define the perimeter. In this study, we implement five distinct boundary conditions (C1 through C5) in \cite{lim2023inverse} governed by boundary parameters ($a,b,d$). Rather than treating all boundary metrics as static values, we allow the parameters to vary dynamically. Modifying these specific parameters significantly alters the resulting electrostatic field distribution, as demonstrated in Figure 1.
\begin{equation} 
\begin{aligned}
\mathbf{x}_{PINN}=
\begin{bmatrix}
\mathrm{a},
\mathrm{b}, 
d^{(t)}
\end{bmatrix}
\in\mathbb{R}^{3 \times 1}.
\\
\mathbf{I}_{PINN} = \left\{ \mathbf{I}^{(1)},\mathbf{I}^{(2)},\dots,\mathbf{I}^{(15)} \right\}
\in\mathbb{R}^{13\times 13\times 15}
\end{aligned}
\end{equation}

\begin{figure} 
   \centering \includegraphics[width=.4\textwidth]{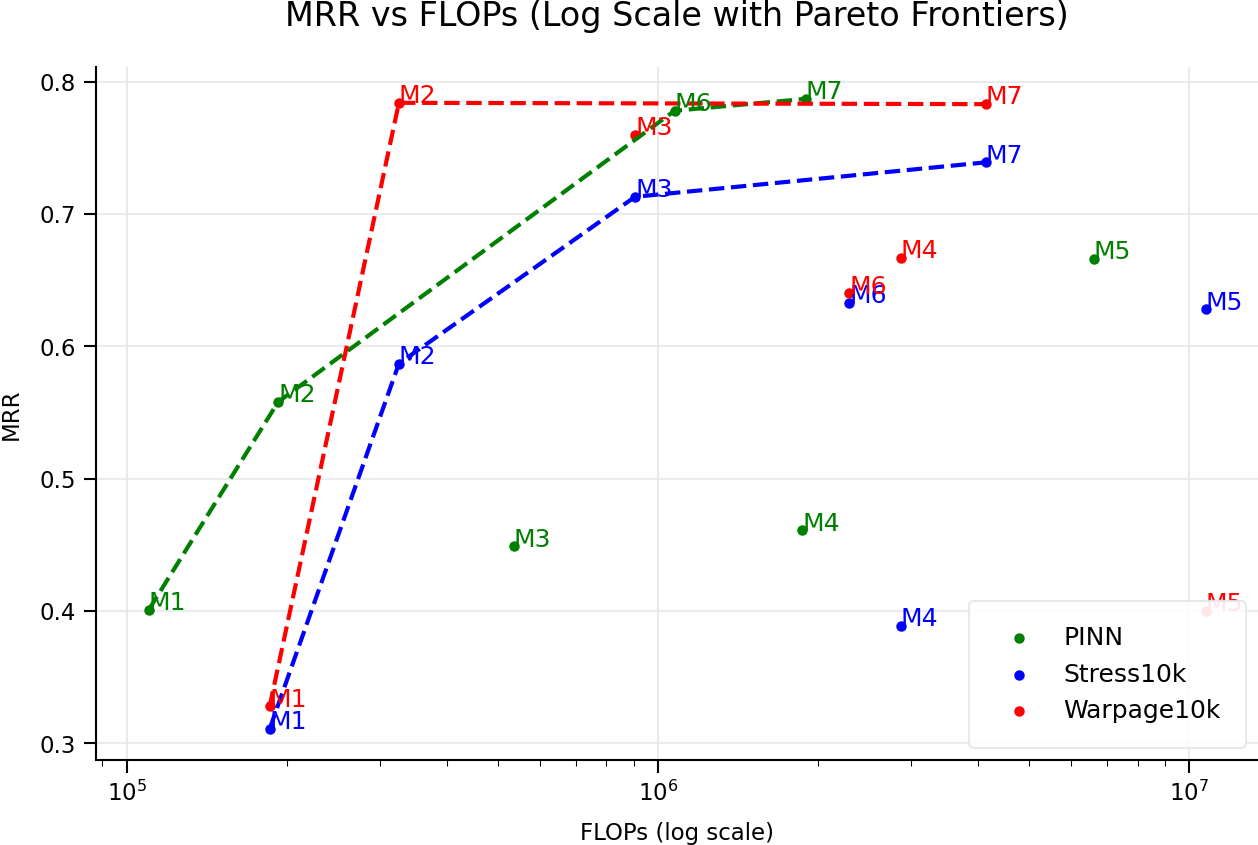} 
    \caption{Higher FLOPs means a GPU takes longer to run on a single request. 
    } \label{fig:pareto_copilot_mmr_flops}
\end{figure}

\begin{figure}
  \centering \includegraphics[width=.4\textwidth]{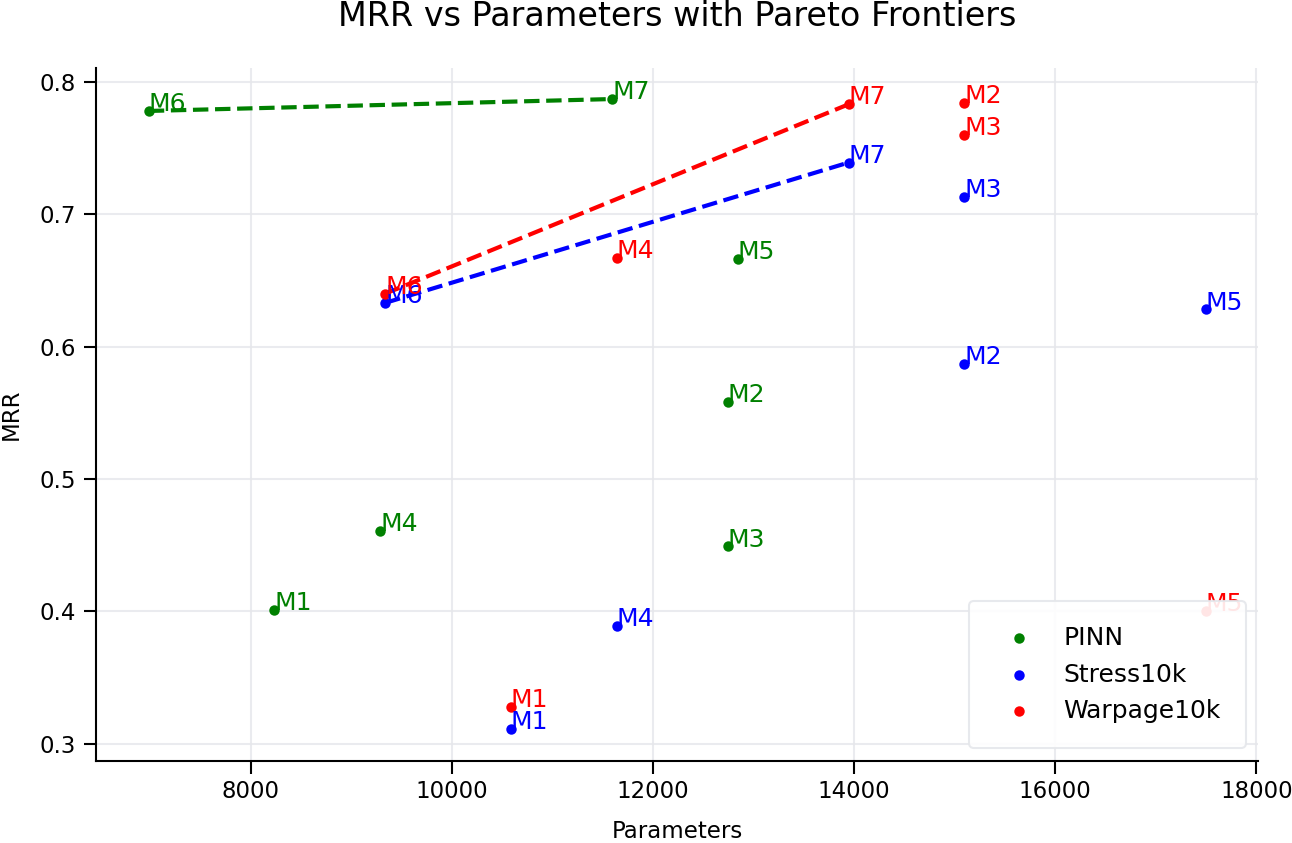} 
    \caption{
    Unlike FLOPs, we cannot "slowly" load a model into VRAM while running it; the entire model must fit into VRAM before inference begins.} \label{fig:pareto_copilot_mmr_params}
\end{figure}

\subsection{Pareto Analysis}
Figure~\ref{fig:pareto_copilot_mmr_flops} compares MRR against FLOPs across the three datasets. The Pareto frontiers differ across datasets: Stress10K progresses from M1$\rightarrow$M2$\rightarrow$M3$\rightarrow$M7, indicating that M3 remains a competitive intermediate operating point; PINN follows M1$\rightarrow$M2$\rightarrow$M6$\rightarrow$M7, where depth conditioning first becomes Pareto-optimal. Warpage10K transitions M2$\rightarrow$M7, suggesting that the intermediate recursive variants are dominated. 
The TRM variants (M4--M5) do not lie on the Pareto frontier for any dataset, indicating a less favorable accuracy-to-computation trade-off.
Figure~\ref{fig:pareto_copilot_mmr_params} compares MRR against Parameters. Unlike the FLOP analysis, the Pareto frontiers consistently favor DEPTH. For PINN, Stress10K and Warpage10K, the Pareto frontier consists exclusively of the DEPTH models (M6--M7), where additional parameters yield further accuracy improvements while remaining Pareto-optimal. The TRM variants (M4--M5) remain well inside the Pareto region despite their larger parameter count, indicating that increasing parameter alone does not translate into better MRR.

\section{Conclusion}
The proposed DEPTH (M6, M7) deliver the strongest or near-strongest retrieval accuracy while using among the fewest parameters and lowest FLOPs of all seven models evaluated. 
The naive weight-sharing recursion of SIMPLE (M2,M3) is FLOP-efficient but accuracy-limited relative to M7 at comparable compute, while the dual-latent-state recursion of TRM (M4,M5) is both parameter- and FLOP-expensive without a commensurate accuracy return. These results support the central claim that explicitly uses depth as a recursively injected input in which each recursive output contributes an individual loss term during BPTT, is the primary factor behind the favorable accuracy-per-parameter and accuracy-per-FLOP trade-offs achieved for package-level Stress10k, Warpage10k, and PINN prediction.
 

\balance
\bibliographystyle{IEEEtran}
 
\bibliography{name}

\end{document}